\documentclass[10pt,twocolumn,letterpaper]{article}

\usepackage{wacv}
\usepackage{times}
\usepackage{epsfig}
\usepackage{graphicx}
\usepackage{amsmath}
\usepackage{amssymb}

\usepackage{gensymb}
\usepackage{lipsum}
\usepackage{float}
\usepackage{stfloats}
\usepackage{multicol}
\usepackage{multirow}
\usepackage{bm}
\usepackage{bbm} 
\usepackage{etoolbox}
\usepackage{array}
\usepackage{tabulary}
\usepackage[table,dvipsnames]{xcolor}
\usepackage{paralist}
\usepackage{booktabs}
\usepackage{adjustbox}
\usepackage{subcaption}
\usepackage[accsupp]{axessibility}

\usepackage{algorithm}
\usepackage{listings}
\makeatletter
\@namedef{ver@everyshi.sty}{}
\makeatother
\usepackage{tikz}
\newcommand*\circled[1]{\tikz[baseline=(char.base)]{
\node[shape=circle,fill=gray,inner sep=0.5pt] (char) {\textcolor{white}{\footnotesize \textbf{#1}}};}}
\newcommand*\circledbrown[1]{\tikz[baseline=(char.base)]{
\node[shape=circle,fill=brown,inner sep=0.5pt] (char) {\textcolor{white}{\footnotesize \textbf{#1}}};}}

\newcommand{\green}[1]{\textcolor[RGB]{96,177,87}{#1}}
\newcommand{\fn}[1]{{#1}}
\newcommand{\gbf}[1]{\green{\bf{\fn{(#1)}}}}

\wacvfinalcopy 

\usepackage[pagebackref=true,citecolor=blue,breaklinks=true,colorlinks,bookmarks=false]{hyperref}

\begin{document}

\title{Trans4Map: Revisiting Holistic Bird's-Eye-View Mapping from Egocentric Images to Allocentric Semantics with Vision Transformers}

\author{Chang Chen,~~Jiaming Zhang\thanks{Correspondence: jiaming.zhang@kit.edu},~~Kailun Yang,~~Kunyu Peng,~~Rainer Stiefelhagen\\
CV:HCI Lab,~~Karlsruhe Institute of Technology
}

\maketitle
\thispagestyle{empty}
\begin{abstract}
Humans have an innate ability to sense their surroundings, as they can extract the spatial representation from the egocentric perception and form an allocentric semantic map via spatial transformation and memory updating. However, endowing mobile agents with such a spatial sensing ability is still a challenge, due to two difficulties: (1) the previous convolutional models are limited by the local receptive field, thus, struggling to capture holistic long-range dependencies during observation; (2) the excessive computational budgets required for success, often lead to a separation of the mapping pipeline into stages, resulting the entire mapping process inefficient. To address these issues, we propose an end-to-end one-stage Transformer-based framework for Mapping, termed Trans4Map. Our egocentric-to-allocentric mapping process includes three steps: (1) the efficient transformer extracts the contextual features from a batch of egocentric images; (2) the proposed Bidirectional Allocentric Memory (BAM) module projects egocentric features into the allocentric memory; (3) the map decoder parses the accumulated memory and predicts the top-down semantic segmentation map. 
In contrast, Trans4Map achieves state-of-the-art results, reducing $67.2\%$ parameters, yet gaining a ${+}3.25\%$ mIoU and a ${+}4.09\%$ mBF1 improvements on the Matterport3D dataset.\footnote[1]{Code at: \url{https://github.com/jamycheung/Trans4Map}.} 
\end{abstract}

\section{Introduction}\label{sec1:intr}
Holistic scene understanding has a crucial role in both indoor and outdoor applications, \textit{e.g.,} autonomous driving~\cite{peng2022mass,yang2021projecting,zhang2022bending,zhou2022cross}, indoor exploration and navigation~\cite{katyal2021high,liu2021hida,ou2022indoor,ramakrishnan2020occupancy}, as well as indoor and outdoor mapping~\cite{chaplot2020learning,chen2021panoramic,georgakis2022learning}. 
These tasks are ordinary for humans with excellent spatial perception ability, as they can continuously extract information from the egocentric perspective, and construct the scene semantic map in the allocentric perspective through the memory and spatial transformation.
\begin{figure}[t]
    \begin{center}
       \includegraphics[width=1.0\linewidth, keepaspectratio]{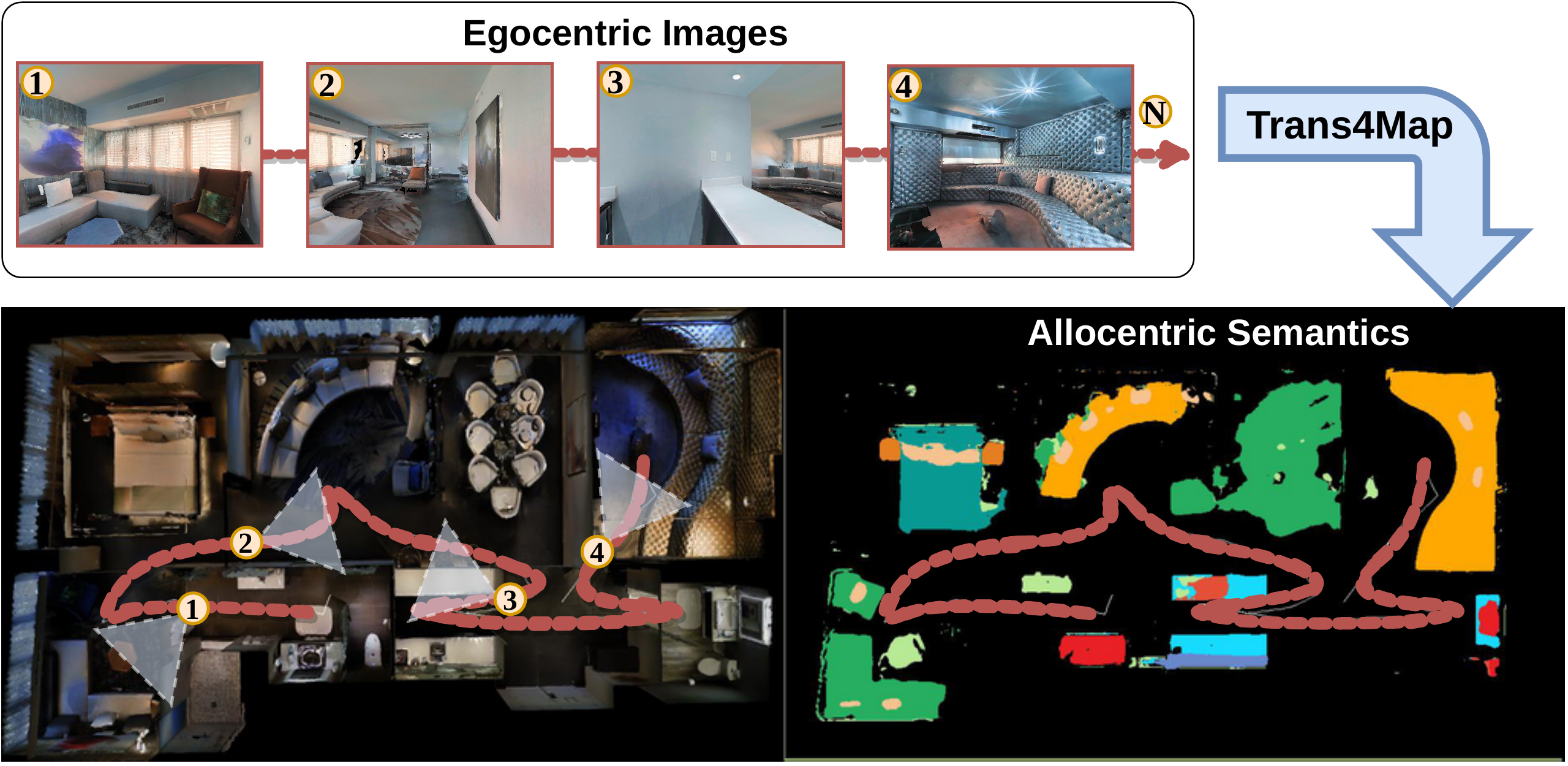}
    \end{center}
        \vskip -2ex
       \caption{ \textbf{The egocentric-to-allocentric semantic mapping.} Given a front-view image sequence of length $N$ observed along a trajectory (the \textcolor[HTML]{B85450}{\textbf{red}} dash line), Trans4Map performs the online extract-project-segment pipeline, yielding an allocentric semantic map in bird's eye view. 
       } 
    \label{fig:main}
\vskip -3ex
\end{figure}

However, the semantic mapping is still difficult for an artificial intelligent mobile agent, particularly when exploring an unfamiliar environment. 
In this work, we focus on the image-based semantic mapping task, by predicting allocentric semantic segmentation from egocentric images.
As the example shown in Fig.~\ref{fig:main}, given a trajectory in the scene, which is composed of a batch of first-view RGB images and the corresponding known camera pose, the mobile agent performs three steps: (1) extracting rich and compact contextual features; (2) projecting and updating the egocentric features in the online intermediate allocentric memory as spatial-semantic representation of the complex spaces; (3) parsing and predicting the final top-view semantic mapping via the decoder. 
The image-based egocentric-to-allocentric mapping pipeline is more in line with human intuition, and is able to perform mapping in an efficient way, avoiding the need for a time-consuming reconstruction phase~\cite{grinvald2019volumetric}. 

\begin{figure*}[t]
    \centering 
    \begin{subfigure}{0.4\textwidth}
      \includegraphics[width=\linewidth]{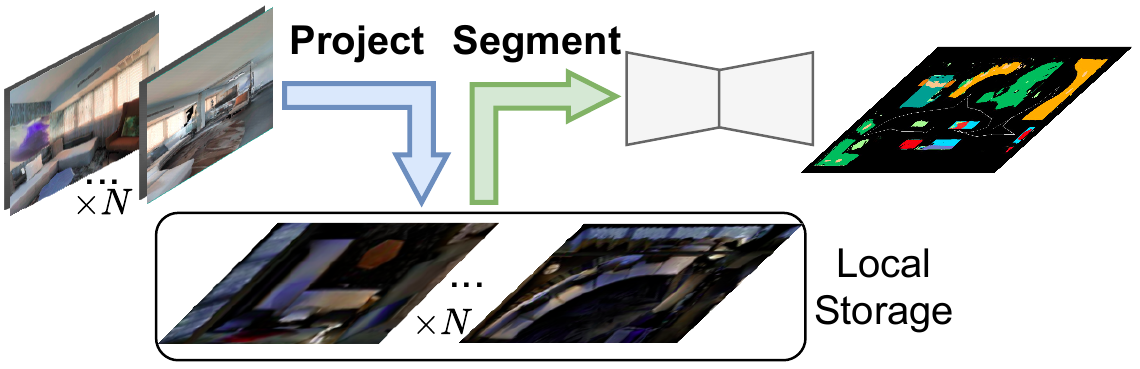}
      \vskip -1ex
      \caption{\footnotesize Two-stage: Project~${\rightarrow}$~Segment}
      \label{fig:pipeline1}
    \end{subfigure}
    \begin{subfigure}{0.4\textwidth}
      \includegraphics[width=\linewidth]{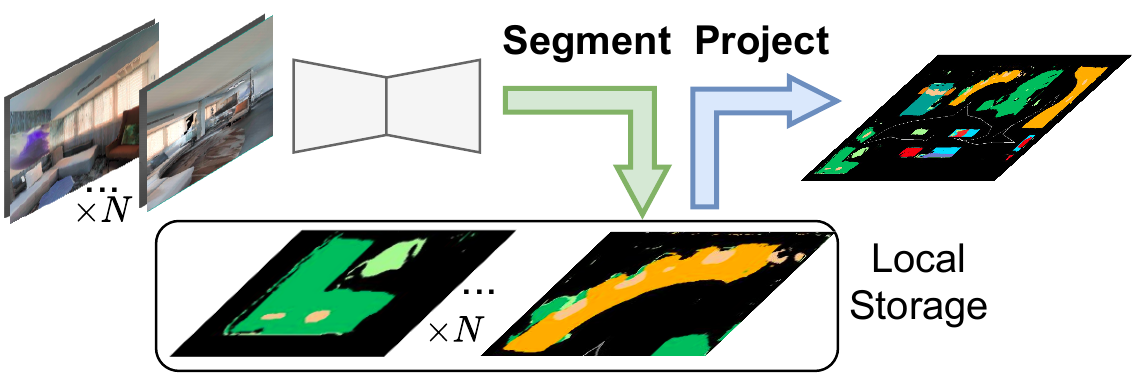}
      \vskip -1ex
      \caption{\footnotesize Two-stage: Segment~${\rightarrow}$~Project}
      \label{fig:pipeline2}
    \end{subfigure} 
    \begin{subfigure}{0.15\textwidth}
      \includegraphics[width=\linewidth]{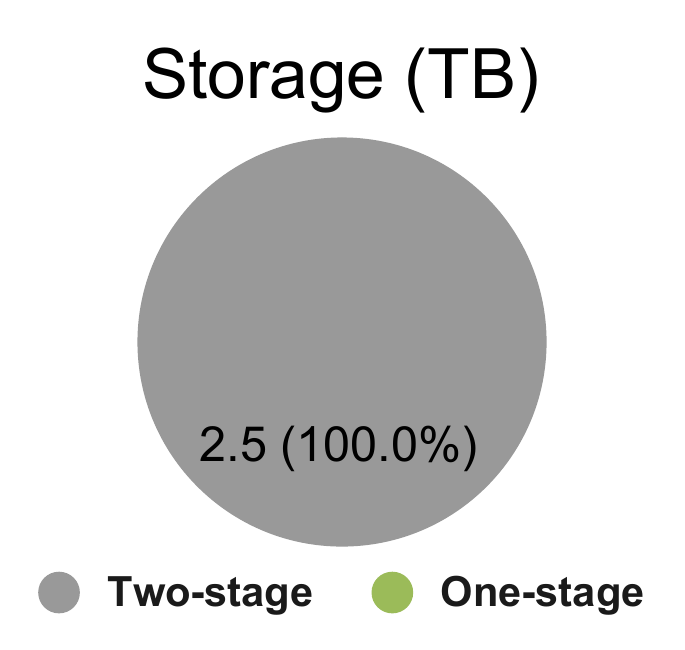}
      \vskip -1ex
      \label{fig:storage}
    \end{subfigure} 
    
    \medskip
    \begin{subfigure}{0.4\textwidth}
      \includegraphics[width=\linewidth]{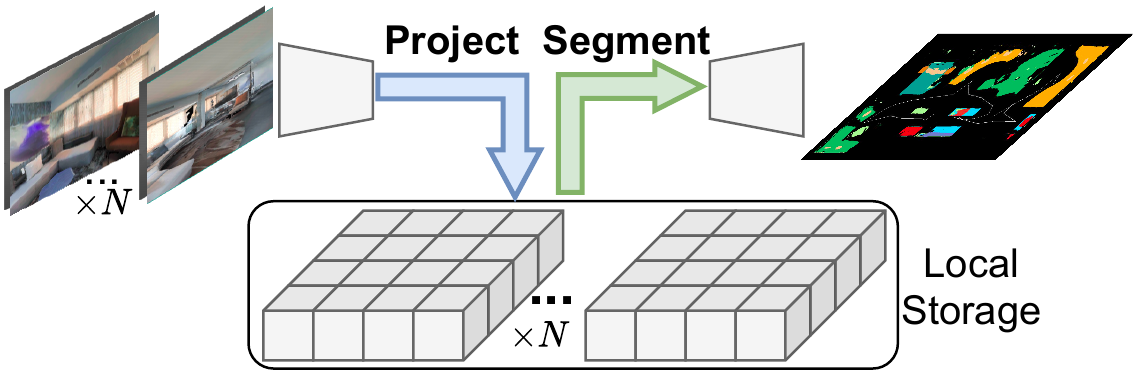}
      \vskip -1ex
      \caption{\footnotesize Two-stage: Offline Project}
      \label{fig:pipeline3}
    \end{subfigure}
    \begin{subfigure}{0.4\textwidth}
      \includegraphics[width=\linewidth]{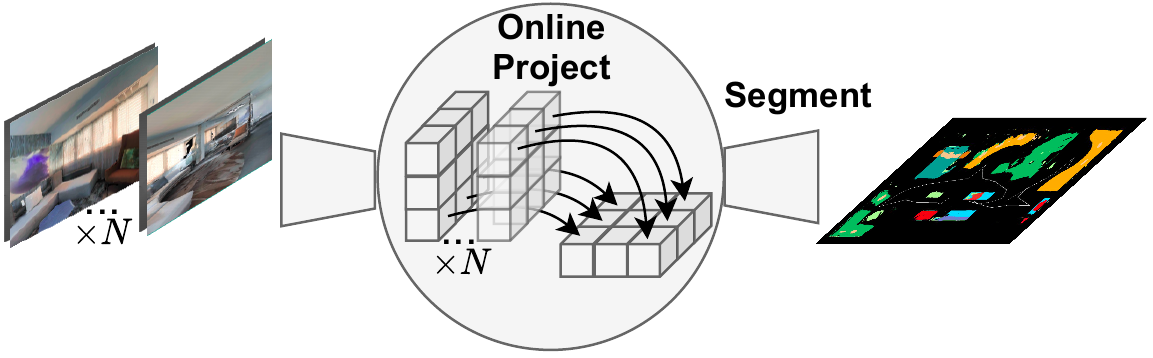}
      \vskip -1ex
      \caption{\footnotesize One-stage: Online Project}
      \label{fig:pipeline4}
    \end{subfigure} 
    \begin{subfigure}{0.15\textwidth}
      \includegraphics[width=\linewidth]{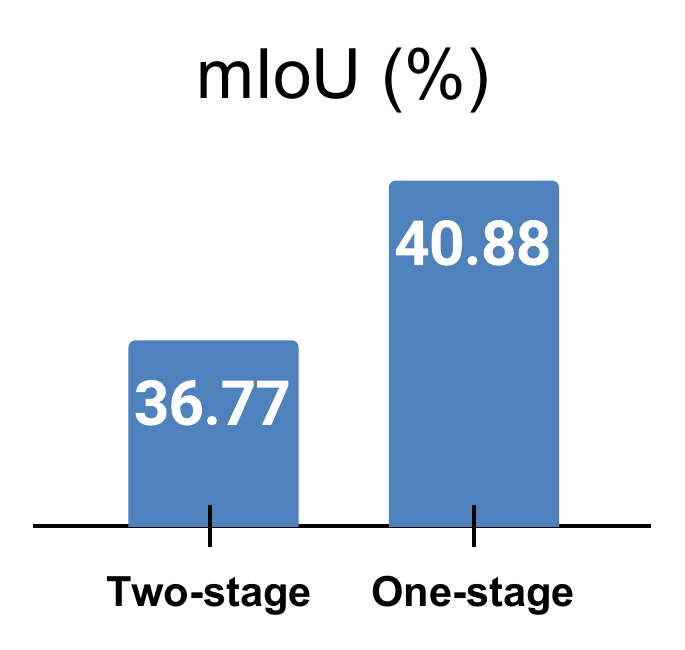}
      \vskip -1ex
      \label{fig:miou}
    \end{subfigure} 
    \vskip -2ex
    \caption{\small \textbf{Semantic mapping pipelines.} The two-stage pipelines (a)(b)(c) differ from the projection locations, \ie, early, late, and intermediate offline projection. The one-stage pipeline (d) avoids $2.5$TB storage by using online projection and has a higher mIoU.}
    \label{fig:pipeline}
\vskip -2ex
\end{figure*}

In Fig.~\ref{fig:pipeline}, image-based semantic mapping methods are divided into four pipelines. The project-then-segment pipeline (Fig.~\ref{fig:pipeline1}) projects $N$ high-resolution observations into Bird's Eye View (BEV), that hinders the small object segmentation due to the lack of fine visual information. The segment-then-project pipeline (Fig.~\ref{fig:pipeline2}) depends heavily on the front-view segmentation performance and may accumulate errors from one to another stage. The offline project-then-segment pipeline (Fig.~\ref{fig:pipeline3}) requires large-scale local storage to save the feature map given by the pre-trained encoder of the first stage. It further demands huge GPU memory to reload offline features for the second stage training. Unlike two-stage pipelines above, our proposed online project-then-segment pipeline (Fig.~\ref{fig:pipeline4}) performs online implicit projection and enables end-to-end and resource-friendly  BEV semantic mapping. The one-stage pipeline is crucial, because it fits resource-limited platforms, \eg, robots. Further, it helps mobile agents to quickly construct maps and get familiar with the unknown space.

However, a lightweight but effective backbone that requires few resources is the decisive factor to achieve successful one-stage semantic mapping. 
The vision transformer architecture~\cite{vaswani2017attention} is able to capture long-distance contextual dependencies, forming a non-local representation. This mechanism naturally fits the semantic mapping task, since the mapping process demands a holistic understanding of scenes. 
This assumption leads us to revisit the top-down semantic mapping with a transformer-based model and put forward a novel end-to-end one-stage \emph{Trans4Map} framework.
It delivers two primary benefits:
(1) the long-range feature modeling ability is advantageous to obtain a more comprehensive spatial representation during the egocentric observation process;
(2) the efficient and lightweight model structure enables the one-stage end-to-end mapping pipeline. Besides, unlike the previous method~\cite{cartillier2020semantic} using a single GRU cell to reload the offline features, we propose a novel \emph{Bidirectional Allocentric Memory (BAM)} to combine features from both directions, which can avoid the occluded objects to be classified as other category, \eg, \emph{chairs} under \emph{tables}. Further, our BAM implicitly performs the efficient online projection, as another key point in implementing the one-stage mapping pipeline (Fig.~\ref{fig:pipeline4}).

\begin{figure}[t]
   \includegraphics[width=1.0\linewidth, keepaspectratio]{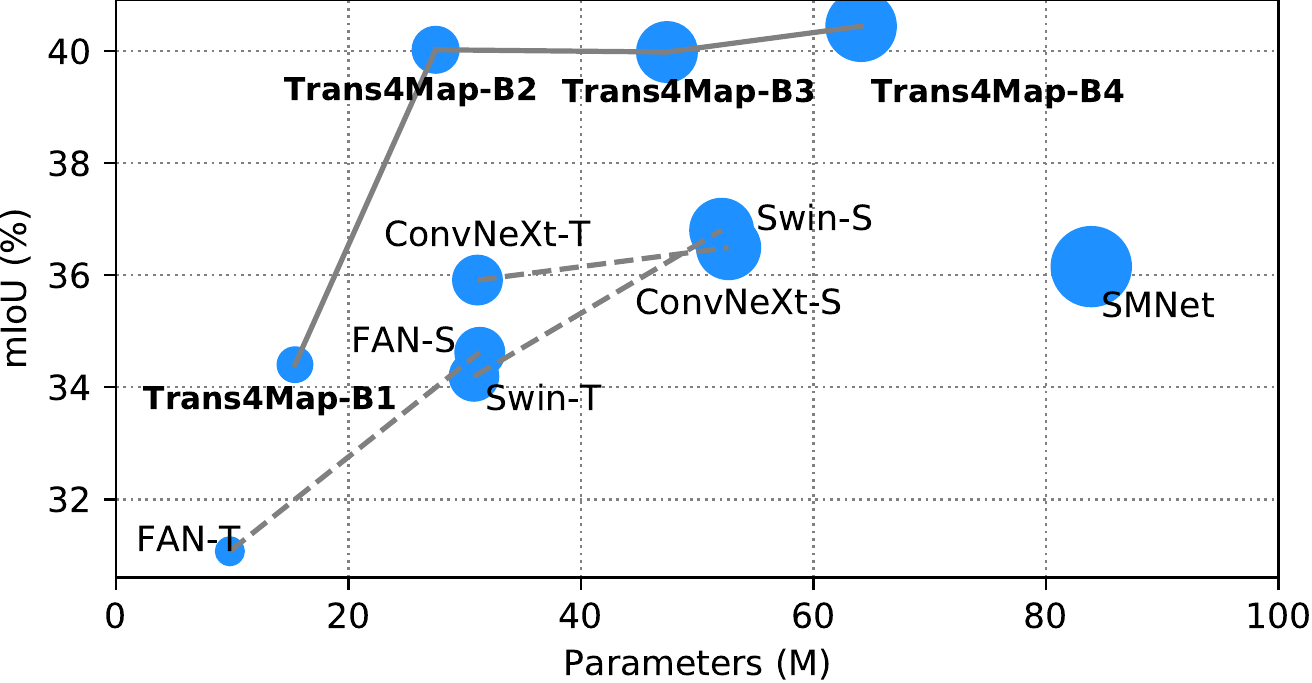}
   \vskip -2ex
   \caption{Semantic mapping scores (mIoU) evaluated by CNN- and Transformer-based backbones with different numbers of parameters (M). Trans4Map models achieve better results yet with fewer parameters. 
   } 
    \label{fig:param_miou}
\vskip -2ex
\end{figure}

To succeed in the end-to-end one-stage Trans4Map framework, we investigate a vast number of advanced deep architectures~\cite{liu2021swin,liu2022convnet,xie2021segformer,zhou2022fan}. According our experiments in Fig.~\ref{fig:param_miou}, we found that simply applying transformer-based backbones does not guarantee improvement.
Thanks to the proposed framework and BAM module, our Trans4Map models have much fewer parameters, yet achieve surprising semantic mapping scores. The B2 version reduces $67.2\%$ parameters compared to SMNet~\cite{cartillier2020semantic} and sets a new state-of-the-art of a ${>}40\%$ mIoU on the Matterport3D~\cite{Matterport3D} dataset.

To summarize, we present the following contributions:
\begin{compactitem}
\item Rethink the top-down semantic mapping task in a one-stage pipeline to fit resource-constrained platforms.
\item Propose an end-to-end \emph{Transformer for Mapping (Trans4Map)} framework to perform egocentric-to-allocentric semantic mapping, yielding a holistic dense understanding for indoor exploration.
\item Put forward a novel \emph{Bidirectional Allocentric Memory (BAM)} which accumulates and projects the egocentric feature to the allocentric spatial tensor, via combining online memories from two directions.
\item Our framework outperforms state-of-the-art counterparts on both the Matterport3D dataset and the Replica dataset, while using a more lightweight model.
\end{compactitem}
\section{Related Work}
\begin{figure*}[t]
    \centering
    \includegraphics[width=1.0\linewidth, keepaspectratio]{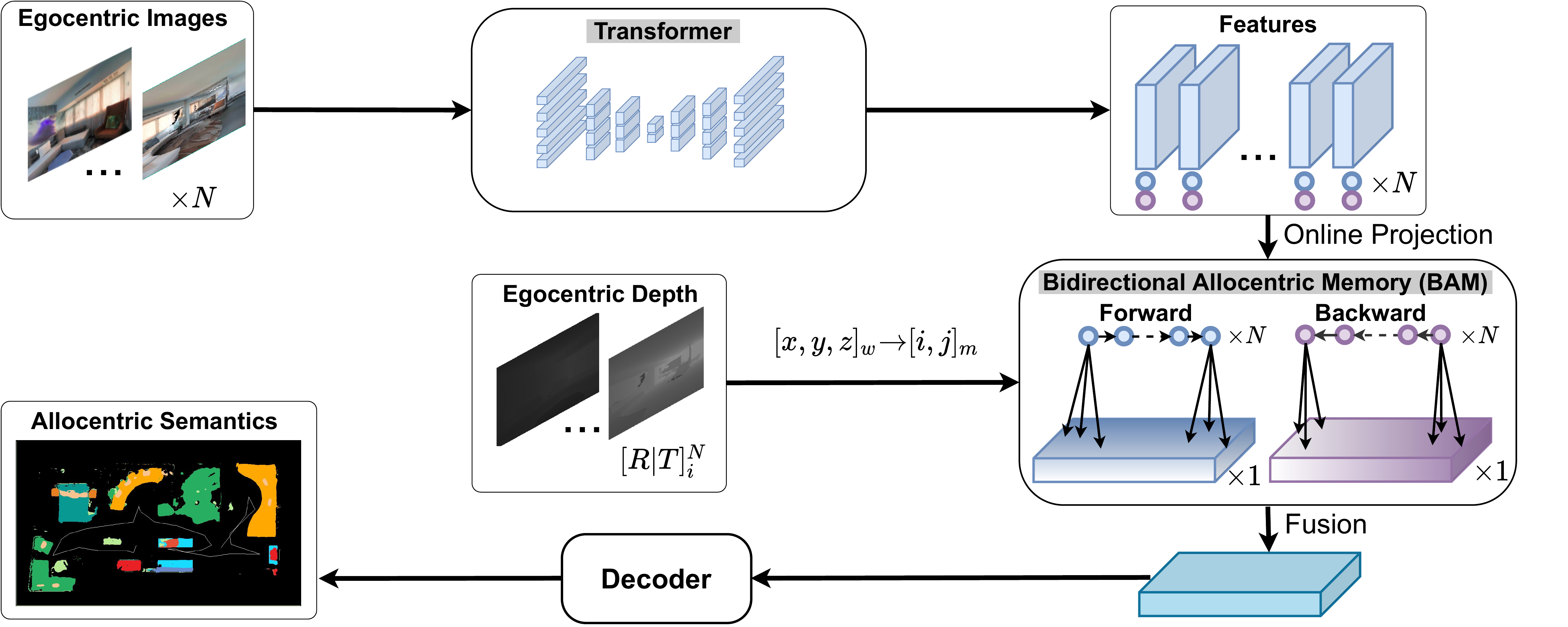}
    \caption{\textbf{The overview of the end-to-end Trans4Map framework.} There are a transformer-based encoder for extracting the egocentric features from the RGB images, a Bidirectional Allocentric Memory (BAM) to project and accumulate the extracted feature sequence to the allocentric feature map via the known depth and pose information, and a CNN-based decoder for parsing the accumulated feature and predicting the allocentric semantics.
    } 
    \label{fig:framework}
    \vskip -2ex
\end{figure*}

\noindent\textbf{Semantic Mapping.}
Recently, a number of approaches have emerged around semantic mapping.
Semantic SLAM pipelines~\cite{grinvald2019volumetric,ran2021rs}
forward the images into a segmentation network and then project predicted labels onto the top-view map. 
These efforts follow the segment-then-project pipeline, which is particularly rigorous for depth information, \textit{i.e.}, the global coordinates of each pixel in RGB images.
Unfortunately, a slight error can lead to an offset in the projection, as well as an under-fitting of model training. 
The project-then-segment pipeline~\cite{singhBMVC18overhead} loses a vast amount of visual information during the projection phase, which hinders the small object segmentation. 
In contrast, SMNet~\cite{cartillier2020semantic} performs offline project-then-segment pipeline, which trains the encoder and decoder separately in two stages and does not optimize the training process from the first-view inputs to the top-view semantics as a whole.

Lu~\textit{et al.}~\cite{lu2019monocular} proposed an end-to-end network encoding the front-view information of the driving scene utilizing a variational encoder-decoder network~\cite{Kingma2014AutoEncodingVB} and then decoding it into a 2D top-down view. Pan~\textit{et al.}~\cite{pan2019crossview} represented a cross-view network with a View Parsing Network (VPN) - an MLP that parses semantics across different views.
Both methods predict a local semantic top-down map with an end-to-end network from egocentric observations. These methods do not encode depth information, so the objects on the semantic map do not reflect their geometry structure.
Moreover, there are many Bird's-Eye-view (BEV) semantic segmentation approaches for driving scene perception~\cite{can2021structured_bev,dosovitskiy2020image,dwivedi2021bird_lifted,gosala2022bird_panoptic,li2022bevformer,saha2021translating,zhao2022scene_representation} emerging in the field.
BEVFormer~\cite{li2022bevformer} aggregates spatiotemporal cues from surround-view cameras, whereas ViT-BEVSeg~\cite{dutta2022vit_bevseg} uses a spatial transformer decoder for generating semantic occupancy grid maps.
Differing from these works, we revisit various backbones including CNN- or Transformer-based and propose a transformer-based end-to-end framework, which acts as a one-stage BEV semantic mapper for holistic \emph{indoor} scene understanding. In addition, the alignment and geometric structure of objects in the generated semantic map are successfully obtained.

\noindent\textbf{Allocentric Spatial Memory.}
Incrementally generating a top-view map from egocentric observations requires dynamically updating the allocentric memory, \ie, aggregating information over time, such as a mobile agent traveling around an indoor scene. Visual SLAM pipelines~\cite{campos2021orb,murTRO2015,thrun2002probabilistic} for this task involve multiple modules such as tracking, mapping, relocalization, loop closure, and graph optimization through bundle adjustment.
MapNet~\cite{henriques2018mapnet} developed RNN to update memory and registers new observations via dense matching.
Tung~\textit{et al.}~\cite{Tung2019LearningSC} proposed Geometry-aware Recurrent Neural Networks~(GRNNs) to segment objects in 3D.
This work is very memory-demanding due to the high-dimension features.
The closest work to our approach is SMNet~\cite{cartillier2020semantic} which uses a GRU to update the projected tensor. 
Unlike previous works, we propose a bidirectional allocentric memory that can better accumulate information over time and segment occluded objects. This crucial design allows Trans4Map to perform implicit online projection, enabling the one-stage mapping pipeline and setting the new state-of-the-art in the large-scale indoor scenes.

\section{Methodology}
In this section, we revisit the allocentric semantic mapping task with vision transformers and introduce our proposed end-to-end one-stage framework, that can generate allocentric semantic maps from egocentric observations. 

\subsection{Trans4Map: Framework Overview}\label{sec3.1framework}
As shown in Fig.~\ref{fig:framework}, our end-to-end Trans4Map framework includes three steps: (1) the incoming $N$ egocentric images are fed into the transformer-based backbone (in Sec.~\ref{sec:transformer_backbone}), which extracts contextual feature and long-range dependency; (2) the Bidirectional Allocentric Memory (BAM) module (in Sec.~\ref{sec:bam}) projects the extracted feature via the depth-based transformation index; (3) the lightweight CNN-based decoder parses the projected feature and predicts the allocentric semantics. 

\noindent\textbf{One-stage pipeline}. An efficient one-stage semantic mapping pipeline is crucial for fast constructing map and is required to deploy mobile agents on resource-limited platforms. The previous SMNet~\cite{cartillier2020semantic} uses a two-branch CNN-based backbone to extract features from RGBD images. Due to the heavy dual-branch backbone, the data pipeline is divided into multiple stages: extracting RGBD feature maps by using a frozen encoder; storing the feature maps locally; and reloading them to fine-tune the decoder.
Unlike such a multi-stage data flow, our framework shown in Fig.~\ref{fig:framework} operates in an one-stage end-to-end manner, benefiting from three designs: (1) a transformer-based backbone is leveraged to capture holistic features and long-range dependencies, instead of narrow-receptive-field CNN-based backbones; (2) a single branch structure for extracting RGB features makes the whole model more lightweight than the dual-branch one; (3) an online training pipeline from egocentric images to allocentric semantics is constructed, avoiding using the time-consuming two-stage process and feature maps storage. Following the one-stage pipeline, our framework can achieve superior allocentric semantic mapping, while maintaining efficiency.

\subsection{Transformer backbone}\label{sec:transformer_backbone}

\begin{figure}[t]
    \centering
    \includegraphics[width=1.0\linewidth, keepaspectratio]{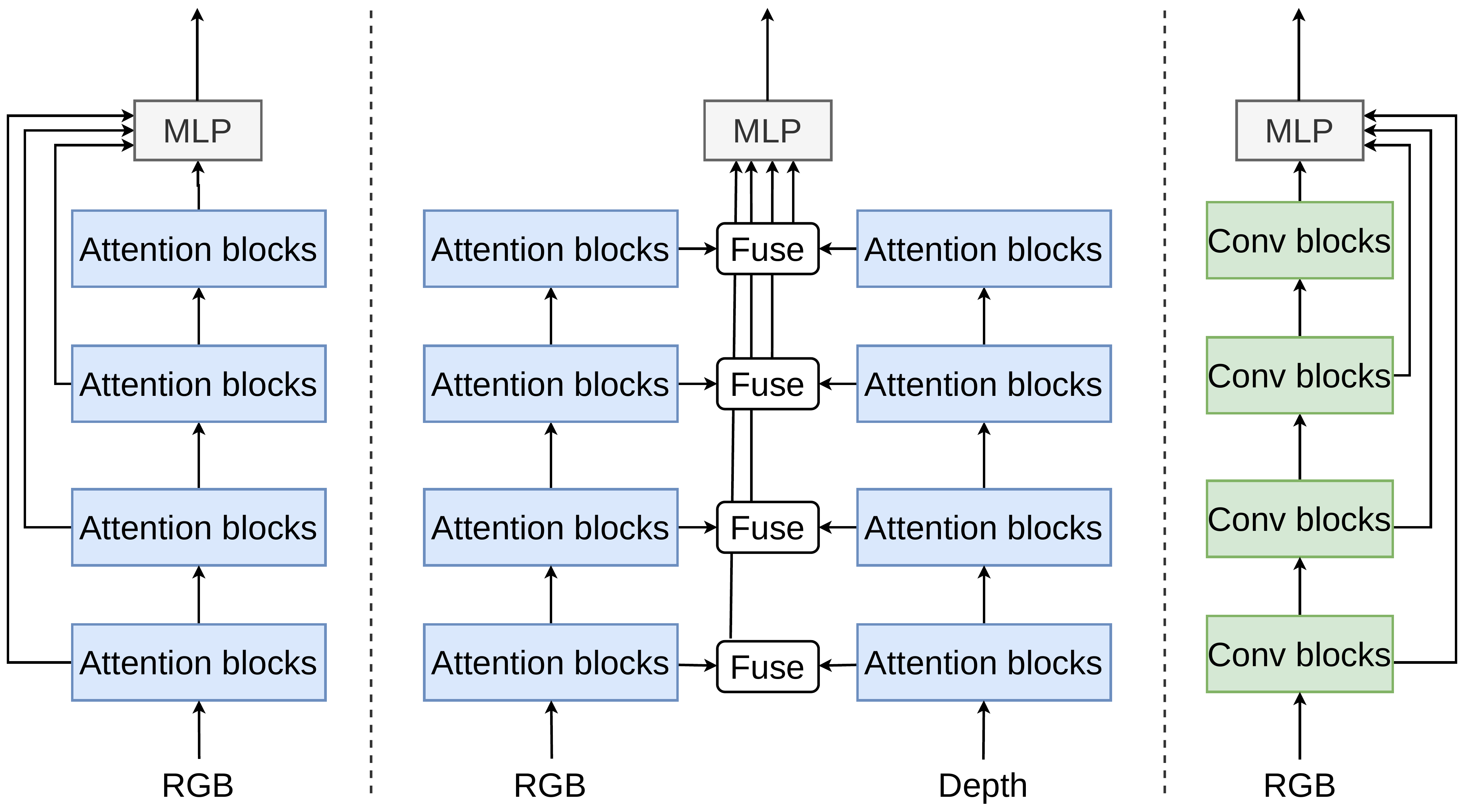}
    \begin{minipage}[t]{.25\linewidth}
    \vskip -3ex
    \subcaption{\small Transformer}\label{fig:encoder1}
    \end{minipage}%
    \begin{minipage}[t]{.55\linewidth}
    \centering
    \vskip -3ex
    \subcaption{\small Multimodal Transformer}\label{fig:encoder2}
    \end{minipage}%
    \begin{minipage}[t]{.2\linewidth}
    \centering
    \vskip -3ex
    \subcaption{\small CNN}\label{fig:encoder3}
    \end{minipage}%
    \vskip -3ex
    \caption{\textbf{Semantic mapping architectures.}} 
    \label{fig:encoder}
    \vskip -3ex
\end{figure}

To fully investigate the proposed Trans4Map framework, we explore different model architectures and learning modalities for the allocentric semantic mapping task, as shown in Fig.~\ref{fig:encoder}. The architectures are constructed by four stages, and each stage includes a series of convolutional blocks (see Fig.~\ref{fig:encoder3}) or self-attention blocks (see Fig.~\ref{fig:encoder1}). Different from the convolutional architecture, the transformer-based architecture is able to capture non-local features, thanks to the self-attention operation~\cite{vaswani2017attention}.
Considering that cross-modality complementary features are informative for predicting semantics~\cite{hu2019acnet,jiang2018rednet,liu2022cmx}, we leverage RGB-Depth inputs and a multimodal architecture (see Fig.~\ref{fig:encoder2}) is reformed by using efficient self-attention blocks.

For brevity, we describe the operation of the single-modal process, while the bimodal process involves an additive fusion at each stage, in which the fusion block obtains the extracted contextual features and geometry features and then fuses them per pixel with the same dimension.
Given a batch of RGB images of size $N{\times}H{\times}W{\times}3$, the divided patches are passed through the four-stage transformer blocks, to obtain the hierarchical feature representation with downsampling rates of $\{\frac{1}{r_1},\frac{1}{r_2},\frac{1}{r_3},\frac{1}{r_4}\}$ and increasing channels of $\{C_1,C_2,C_3,C_4\}$.
Then, the multi-scale features are concatenated by an MLP layer and followed by a convolution layer with $64$ channels. So the hierarchical features are fused into an egocentric feature of size $N{\times}\frac{H}{r_1}{\times}\frac{W}{r_1}{\times}64$.
To study different semantic mapping architectures, the multi-scale features in this work are extracted with $\{\frac{1}{4}, \frac{1}{8}, \frac{1}{16}, \frac{1}{32}\}$ downsampling rates and $\{64, 128, 320, 512\}$ channels.

To compare CNN- and transformer-based models, the RedNet backbone~\cite{jiang2018rednet} used in SMNet and the ConvNeXt backbone are selected to form CNN-based mapping models, while transformer-based models include FAN~\cite{zhou2022fan}, Swin~\cite{liu2021swin}, and SegFormer~\cite{xie2021segformer} backbones.
Based on our experiments, we adopt SegFormer~\cite{xie2021segformer} as the default backbone of our visual encoder, as its simple and lightweight design can generate features ranging from high-resolution fine features to low-resolution coarse features.
More ablation studies and discussions are unfolded in Sec.~\ref{exp:ablation}.

\subsection{Bidirectional Allocentric Memory}\label{sec:bam}
After acquiring the egocentric features through the aforementioned transformer backbone, the projective index is needed to project representative contextual features into an allocentric memory map.
In the Habitat simulator~\cite{habitat19iccv}, we can directly obtain the state of the moving agent and then calculate the camera pose using relative orientation and position. 
In order to perform the online projection, we need to derive the 3D position of each pixel in the egocentric image, as presented in Eq.~\eqref{eq:eq1} and Eq.~\eqref{eq:eq2}.

\begin{equation}
\label{eq:eq1}
{
\left[ \begin{array}{ccc}
x\\
y\\
z
\end{array} 
\right ]_c = K^{-1} 
\left[ \begin{array}{ccc}
u\\
v\\
d_{u,v}
\end{array} 
\right ]_i
}
\end{equation}
\begin{equation}
\label{eq:eq2}
{
\left [ \begin{array}{ccc}
X\\
Y\\
Z
\end{array} 
\right ]_w = R^{-1} 
\left[ \begin{array}{ccc}
x\\
y\\
z
\end{array} 
\right ]_c - \overrightarrow{t}
}
\end{equation}

$K$ in Eq.~\eqref{eq:eq1} is the camera intrinsic parameter matrix, $[R | \overrightarrow{t}]$ in Eq.~\eqref{eq:eq2} are the rotation matrix and the translation matrix, respectively.
First, in Eq.~\eqref{eq:eq1}, using the pinhole camera model and the depth of each pixel $d_{u,v}$, the pixel coordinate $(u,v)$ in the image coordinate system can be converted into the camera coordinate system.
Then, in Eq.~\eqref{eq:eq2}, the camera coordinates denoted as $(x, y, z)$ of each point are converted to world coordinates denoted as $(X, Y, Z)$ using the rotation matrix and the translation matrix.

Each pixel in the allocentric memory map represents a ${2cm\times2cm}$ cell in the scene of the Matterport3D dataset~\cite{Matterport3D}, so the projective index $(i,j)_m$ can be calculated by dividing the world coordinates $X$ and $Z$ of each point by the resolution. Finally, we project egocentric features of $N$ batch size onto the allocentric memory map using the calculated projective index. 

\begin{figure}[t]
    \centering
    \includegraphics[width=1.0\linewidth, keepaspectratio]{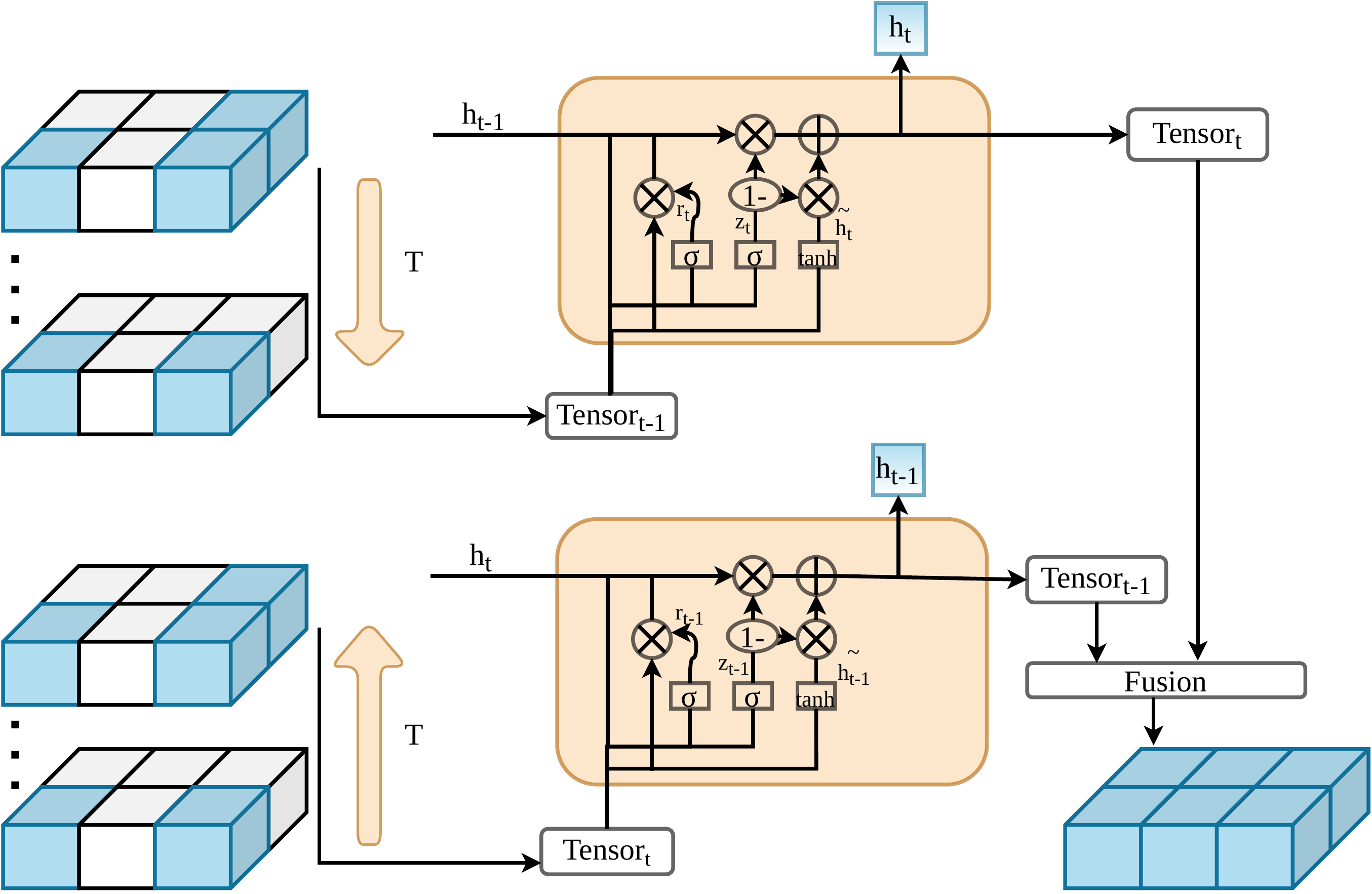}
    \vskip -2ex
    \caption{\textbf{Bidirectional Allocentric Memory (BAM).}} 
    \label{fig:bi_gru}
    \vskip -2ex
\end{figure}

To enhance the long-range content dependency and aggregate the incoming information completely, we propose Bidirectional Allocentric Memory ~(\emph{BAM}), in which we transmit projected features via Bi-directional GRU~(BiGRU), which are used to update and accumulate incoming observations from two directions. Specifically, BAM adds a reverse GRU unit that performs feature parsing and accumulation.
As shown in Fig.~\ref{fig:bi_gru}, the upper GRU unit processes the allocentric memory tensor in a forward direction from the $M^{t-1}$ to the $M^{t}$ feature and the lower GRU unit in a backward direction from the $M^{t}$ to the $M^{t-1}$ feature. A simple yet effective convolutional layer is applied to fuse two projected allocentric memory features.
The computation of updated spatial memory tensor is formulated as:
\begin{equation}
    M^t_{i,j} = GRU(F^t_{i,j},M^{t-1}_{i,j});
    \label{Forword}
\end{equation}
\begin{equation}
    M^{t-1}_{i,j} = GRU(F^{t-1}_{i,j},M^t_{i,j});
    \label{Backword}
\end{equation}
\begin{equation}
    T = Conv(M^t_{i,j}, M^{t-1}_{i,j}).
    \label{}
\end{equation}
$M^t_{i,j}$ and $M^{t-1}_{i,j}$ are the current time step spatial memory and previous time step spatial memory, respectively. The fused spatial memory tensors $T$ are accessible to the decoding step for the final semantic top-down map prediction.
Thanks to the bidirectional parsing process, BAM is able to accumulate the observations per each time step in both direction in parallel, thus, it can better avoid occluded objects being wrongly-classified. With BAM, Trans4Map can produce a more meaningful allocentric representation which combines bidirectional projected features.

\section{Experiment}

\subsection{Datasets}
\noindent \textbf{Matterport3D}. The Matterport3D dataset~\cite{Matterport3D} contains photo-realistic scans of $90$ building-scale environments. It provides RGBD images and 3D annotations with $40$ categories, which are crucial for scene understanding tasks. We follow the same dataset split as SMNet~\cite{cartillier2020semantic}, consisting of $61$ training scenes, $7$ validation scenes, and $17$ test scenes, which mainly focuses on $12$ object categories: \emph{chair, table, cushion, cabinet, shelving, sink, dresser, plant, bed, sofa, counter, fireplace}. Other rare objects and floor surfaces are masked as void classes.
Given a trajectory, we sample $50$ sets of $N$ consecutive waypoints in each unique scene via Habitat simulator~\cite{habitat19iccv}. The $N$ consecutive RGBD image sequences as a batch are forwarded into the model as input. 

\noindent \textbf{Replica}. The Replica dataset~\cite{straub2019replica} contains $18$ various highly photo-realistic indoor environments. It provides dense-mesh, high-resolution RGBD images and also a large range of instance annotations with $88$ categories. We follow the same setup as in the Matterport3D dataset, \textit{i.e.}, focusing on $12$ object categories. We test our model on the Replica using the weight that was trained on Matterport3D dataset, so the $18$ scenes are included in the test split.

\subsection{Implementation details}\label{sec4.2impl}
Performing the online project-segment paradigm, we train our model in an end-to-end manner.
For the ablation study, we revisit the transformer-based mapping and investigate several visual encoders in our model which are pretrained on the ImageNet~\cite{russakovsky2015imagenet} and ADE20K~\cite{ade20k}, and the BAM module and the decoder are randomly initialized.
We use the AdamW optimizer~\cite{kingma2014adam} to train our model with four 1080Ti GPUs on the Matterport3D dataset. The learning rate is initialized with $6e^{-5}$ and then scheduled by LambdaLR. We use Cross-Entropy as the loss function. Training with $100$ epochs will take about $30$ hours. Similar to \cite{cartillier2020semantic}, the evaluation metrics include the pixel-wise accuracy~(Acc), the pixel recall~(mRecall) and precision~(mPrecision) scores, the intersection-over-union~(mIoU) score, and the boundary F1~(mBF1) score~\cite{csurka2013good}.

\begin{table}[t]
\centering
\caption{\textbf{Allocentric semantic mapping results} on the Matterport3D dataset. Model$\dag$ is our implementation.
}
\vskip -2ex
\setlength{\tabcolsep}{1pt}
\resizebox{\columnwidth}{!}{
\begin{tabular}{ l | c c c| c c }
\toprule
\textbf{Method} & \textbf{Acc} & \textbf{mRecall} & \textbf{mPrecision} & \textbf{mIoU} & \textbf{mBF1}\\ 
\midrule\midrule
Seg. GT $\rightarrow$ Proj. & 89.49  & 73.73  & 74.58  & 59.73  & 54.05  \\ \midrule
Two-stage Proj. $\rightarrow$ Seg. & 83.18  & 27.32  & 35.30  & 19.96  & 17.33 \\
Two-stage Seg. $\rightarrow$ Proj. & 88.06  & 40.53  & \textbf{58.92} & 32.76  & 33.21  \\
Two-stage Semantic SLAM & 85.17  & 37.51  & 51.54  & 28.11  & 31.05  \\
Two-stage SMNet & 88.14 & 47.49 & 58.27  & 36.77 & 37.02\\\midrule
Two-stage SMNet~\dag & \textbf{89.14}&46.34&56.98&36.16&35.95 \\
One-stage Trans4Map & 89.02 & \textbf{54.50} & 56.20 & \textbf{40.02} & \textbf{41.11}\\ \bottomrule
\end{tabular}}
\label{tab:exp_mp3d}
\end{table}
 
\begin{table}[t]
\centering
\caption{\textbf{Allocentric semantic mapping results} on the Replica dataset. Model$\dag$ is our implementation. Note that the last two rows are evaluated on the partially available Replica~\cite{straub2019replica} dataset, while the others have all data as~\cite{cartillier2020semantic}.
}
\vskip -2ex
\setlength{\tabcolsep}{1pt}
\resizebox{\columnwidth}{!}{
\begin{tabular}{ l | c c c| c c }
\toprule
\textbf{Method} & \textbf{Acc} & \textbf{mRecall} & \textbf{mPrecision} & \textbf{mIoU} & \textbf{mBF1}\\  \midrule\midrule
Seg. GT $\rightarrow$ Proj. & 96.83  & 83.84  & 94.05  & 79.76  & 86.89  \\ \midrule
Two-stage Seg. $\rightarrow$ Proj. & 88.61  & 48.11  & 65.20 & 40.77  & 45.86 \\
Two-stage Semantic SLAM & 88.30  & 45.80  & 62.41  & 37.99  & 46.71 \\
Two-stage SMNet & 89.26 & 53.37 & 64.81  & 43.12 & 45.18 \\ \midrule
Two-stage SMNet~\dag & \textbf{87.69} & 58.88 & 34.85 & 27.68 & 42.67\\ 
One-stage Trans4Map & 86.19 & \textbf{65.27} & \textbf{34.91} & \textbf{29.15} & \textbf{48.66} \\
\bottomrule
\end{tabular}}
\label{tab:exp_replica}
\end{table}
\subsection{Allocentric semantic mapping results}
As shown in Table~\ref{tab:exp_mp3d} and Table~\ref{tab:exp_replica}, a set of results are conducted on the Matterport3D and the Replica datasets. The experiments assess four pipelines discussed in Fig.~\ref{fig:pipeline}, \ie, two-stage project-then-segment, segment-then-project, offline project, and our one-stage online Trans4Map.

\noindent \textbf{Matterport3D}. 
The results on Matterport3D are in Table~\ref{tab:exp_mp3d}.
Following the segment-project paradigm, the result obtained by using the label data is the upper bound performance.
As in Table~\ref{tab:exp_mp3d}, the segment-project baseline performs much better than the project-segment one, since part of information will be lost in the process of converting an egocentric image into a top-down view.
The semantic SLAM in~\cite{grinvald2019volumetric} also uses the segment-project method but achieves worse performance than the image-based segment-project baseline.
SMNet~\cite{cartillier2020semantic} follows the offline project-segment paradigm and adds a spatial memory update module.
Here, we reproduce the experiment using the released code under the same condition and obtain the results with a mIoU score of $36.16\%$ and a mBF1 value of $35.95\%$. Compared to SMNet, our Trans4Map model achieves significant improvements in terms of mIoU ($40.02\%$) and mBF1 ($41.11\%$) on the Matterport3D dataset, which proves the effectiveness of our proposed allocentric mapping framework.

\noindent \textbf{Replica}.
The results on Replica are in Table~\ref{tab:exp_replica}. All models are trained on the Matterport3D dataset and tested on the Replica dataset. Note that the trajectories and labels of the Replica dataset are partially available at the moment, thus, the results are tested on the constrained data of the Replica dataset. Nonetheless, under the same condition with the same label data, our Trans4Map outperforms the baseline SMNet with a $1.47\%$ mIoU and a $5.99\%$ mBF1 improvements, respectively. The results indicate that our Trans4Map framework achieves consistent improvements across different datasets.
\begin{table}[t]
    \centering
    \caption{\textbf{Comparison of training resources} between two-stage and one-stage semantic mapping pipeline. The metrics include the local storage (TB), one epoch training time (h:hour), one epoch data loading time (h:hour), RAM requirement (GB), \#Param (M), and mIoU (\%).}
    \vskip -2ex
    \setlength\tabcolsep{1.0pt}
    \resizebox{\columnwidth}{!}{
    \begin{tabular}{l|rrrrc|r}
    \toprule
    Method & Storage (\textbf{TB}) & Train (\textbf{h}) & Load (\textbf{h}) & RAM (\textbf{GB}) & \#Param (\textbf{M}) & mIoU (\textbf{\%})\\ \midrule
    Two-stage & 2.5 & 6.00 & 2.00 & 256 & 83.9 & 36.77\\
    One-stage & \textbf{0} & \textbf{0.33} & \textbf{0.01} & \textbf{18} & \textbf{27.5} & \textbf{40.88} \\
    \rowcolor{gray!15}Change & -100\% & -94.5\% & -99.5\% & -93.0\% & -67.2\% & +11.2\% \\
    \bottomrule
    \end{tabular}}
    \label{tab:train_compare}
\end{table}

\begin{table}[t]
\centering
\caption{\textbf{Comparison of different backbones} including the number of parameters and performance on Matterport3D. Except for two-stage SMNet, all methods are using one-stage Trans4Map framework, so as to ablate the backbone.}
\vskip -2ex
\renewcommand\arraystretch{0.8}
\resizebox{\columnwidth}{!}{
\begin{tabular}{llll}
\toprule
\textbf{Method} & \textbf{Backbone} & \textbf{\#Param (M)} & \textbf{mIoU (\%)}  \\ \midrule\midrule
SMNet & RedNet~\cite{cartillier2020semantic} & 83.9 & 36.77 \\\midrule
ConvNeXt & ConvNeXt-T~\cite{liu2022convnet} & 31.1 & 35.91\\
ConvNeXt & ConvNeXt-S~\cite{liu2022convnet} &  52.7 & 36.49 \\ 
FAN & FAN-T~\cite{zhou2022fan} & 09.8 & 31.07 \\
FAN & FAN-S~\cite{zhou2022fan} & 31.3 & 34.62  \\
Swin & Swin-T~\cite{liu2021swin} & 30.8 & 34.19 \\
Swin & Swin-S~\cite{liu2021swin} & 52.1 & 36.80  \\
Trans4Map & MiT-B1~\cite{xie2021segformer}  & 15.4 & 34.38 \\
Trans4Map & MiT-B2~\cite{xie2021segformer}  & 27.5~\gbf{-56.4} & 40.02~\gbf{+3.25} \\
Trans4Map & MiT-B3~\cite{xie2021segformer}  & 47.4~\gbf{-36.5} & 39.98~\gbf{+3.21} \\
Trans4Map & MiT-B4~\cite{xie2021segformer}  & 64.1~\gbf{-19.8} & \textbf{40.88}~\gbf{+4.11} \\
\bottomrule
\end{tabular}
}
\label{tab:abl_backbone}
\end{table}

\subsection{Ablation study}\label{exp:ablation}
\noindent \textbf{Analysis of One-stage Pipeline}. 
To analyze the efficiency of different semantic mapping pipelines, we present the training process of the two-stage~\cite{cartillier2020semantic} and our one-stage pipeline in Table~\ref{tab:train_compare}. The two-stage method saves the intermediate feature maps offline and reloads them for a second-stage fine-tuning. Compared to the two-stage method, our one-stage method performing online mapping does not require extra local storage ($0$TB~\emph{vs.}~$2.5$TB).
Thanks to our efficient transformer-based model, the one-stage pipeline achieves faster training ($0.33$h~\emph{vs.}~$6$h) and loading ($0.01$h~\emph{vs.}~$2$h) processes than the two-stage one. Further, our RAM requirement ($18$GB~\emph{vs.}~$256$GB) and \#Param ($27.5$M~\emph{vs.}~$83.9$M) are much lower, which is crucial for resource-limited mobile platforms. Surprisingly, the one-stage pipeline surpasses the two-stage one with ${+}4.11\%$ mIoU gains. There are sufficient improvements while saving resources, demonstrating the effectiveness of our proposed one-stage semantic mapping pipeline.

\noindent \textbf{Analysis of Encoder}.
Based on the one-stage pipeline, we analyze the effects of using CNN-based or transformer-based backbones to extract features. 
Table~\ref{tab:abl_backbone} presents the comparison between the two-stage SMNet and the other methods that are all based on the one-stage Trans4Map framework. The baseline SMNet using CNN-based backbone has the largest number of parameters but achieves only $36.77\%$ in mIoU.
We found that simply applying an advanced CNN-based backbone (ConvNeXt~\cite{liu2022convnet}) or transformer-based backbones (FAN~\cite{zhou2022fan} and Swin~\cite{liu2021swin}) does not lead to sufficient improvement. When their backbones are much more lightweight, their performances are even a bit inferior to SMNet.
In contrast, our Trans4Map equipped with the MiT-B2 backbone~\cite{xie2021segformer} reduces $67.2\%$ parameters compared to SMNet, \ie, from $83.9$M to $27.5$M, but the performance has made a sufficient leap in mIoU~(${+}3.25\%$). The MiT-B3 and -B4 backbone with $47.4$M and $64.1$M parameters bring similar mIoU gains (${+}3.21\%$ and ${+}4.11\%$) as the MiT-B2 backbone, because a larger transformer model often requires more training data to obtain a desired boost. 
The results of our proposed Trans4Map confirm that transformer-based models fit the semantic mapping task and can maintain the desired performance with a good trade-off of accuracy and efficiency.

\begin{table}[t]
\centering
\caption{Ablation study of the BAM in Trans4Map. Methods are based on Mit-B4 and four sampling points.}
\renewcommand\arraystretch{0.8}
\begin{tabular}{cll}
\toprule
&\textbf{Method}  & \textbf{mIoU} \\ \midrule\midrule
& GRU (SMNet~\cite{cartillier2020semantic}) & 36.77\\\midrule
\circled{1} &  GRU + Conv & 37.86\\
\circled{2} &  GRU + 2 GRUCells & 37.67 \\
\circled{3} &  BiGRU + Concatenate & 36.73\\ 
\circled{4} &  BiGRU + gMLP~\cite{gmlp} + Concatenate & 37.49 \\ 
\circled{5} &  BiGRU + gMLP~\cite{gmlp} + Conv Fusion & 40.15\\ 
\circled{6} &  BiGRU + 2 GRUCells + Conv Fusion & 40.00\\ 
\circled{7} &  BiGRU + Conv Fusion (our BAM) & \textbf{40.44}\\ 
\bottomrule
\end{tabular}
\vskip -2ex
\label{tab:abl_gru}
\end{table}

\noindent \textbf{Analysis of BAM}.
Apart from the encoder, we further analyze different structures of our \emph{BAM} module. 
As shown in Table~\ref{tab:abl_gru}, compared the baseline GRU in SMNet and \circled{1}\circled{2}, simply stacking more GRU cells or convolutional layers does not improve model performance. Our proposed convolutional fusion is better than the concatenation fusion, yielding ${+}2.77\%$ gains (\circled{3}~$\rightarrow$~\circled{7}) and ${+}2.66\%$ gains (\circled{4}~$\rightarrow$~\circled{5}), respectively.
The method (\circled{5}) is to follow an advanced gMLP block~\cite{gmlp} for token information mixing after the BiGRU module and yields a mIoU of $40.15\%$. 
The method (\circled{6}) equipped with two GRU cells obtains $40.00\%$ in mIoU. 
Our \emph{BAM} (\circled{7}) applies BiGRU and a convolutional layer to process the spatial tensors and achieves the best performance with $40.44\%$ in mIoU.  The ablation study demonstrates that our BAM is crucial for semantic mapping.

\noindent\textbf{Analysis of pre-train sources.}
Further, we ablate three different pre-train sources.
Comparing \circledbrown{2} and \circledbrown{3} in Table~\ref{tab:abl_pretrain}, Trans4Map model benefits more from the SUN-RGBD dataset than from the ADE20K dataset, since the ADE20K weights are shared between RGB and depth branches, while the SUN-RGBD ones are separated. Between \circledbrown{1} and \circledbrown{4}, both Trans4Map models pre-trained on ImageNet and ADE20K have comparable results. It indicates that our single-modal Trans4Map is stable and robust across different pre-training settings.

\begin{figure*}[t]
    \centering
    \includegraphics[width=1.0\linewidth, keepaspectratio]{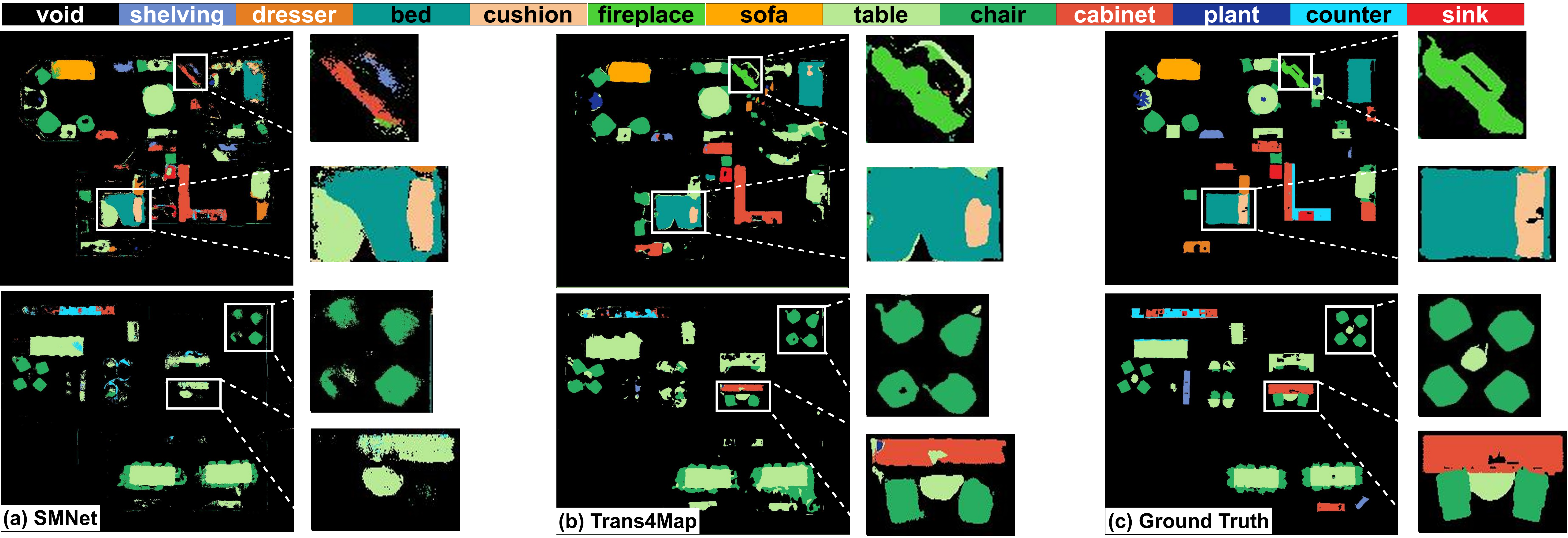}
    \vskip -1ex
    \caption{\textbf{Allocentric semantic mapping visualizations.} There are two indoor scenes from the Matterport3D test set. From left to right are the predicted results of SMNet, the results of our Trans4Map and the ground truth. Zoom in for better view.} 
    \label{fig:vis}
\end{figure*}
\begin{table}[t]
\centering
\caption{Analysis of model complexities, data sources, sampling points in a trajectory, and data modalities.}
\renewcommand\arraystretch{0.8}
\resizebox{\columnwidth}{!}{
\begin{tabular}{clllll}
\toprule
&\textbf{Method} & \textbf{Pre-train} & \textbf{\#Points} & \textbf{Modality} & \textbf{mIoU} \\ \midrule\midrule
&SMNet~\cite{cartillier2020semantic} & SUN-RGBD & 20 & RGBD & 36.77 \\ \midrule
\circledbrown{1} & B2 & ImageNet & 4 & RGB & 37.86\\
\circledbrown{2} & B2 & SUN-RGBD & 4 & RGBD & 40.27\\
\circledbrown{3} & B2 & ADE20K & 4 & RGBD & 40.15\\
\circledbrown{4} & B2 & ADE20K & 4 & RGB & 37.71 \\
\circledbrown{5} & B2 & ADE20K & 20 & RGB & 40.02\\
\circledbrown{6} & B3 & ADE20K & 4 & RGB & 38.78\\
\circledbrown{7} & B3 & ADE20K & 20 & RGB & 39.98\\
\circledbrown{8} & B4 & ADE20K & 4 & RGB & 40.44\\
\circledbrown{9} & B4 & ADE20K & 20 & RGB & \textbf{40.88}\\
\bottomrule
\end{tabular}
}
\vskip -2ex
\label{tab:abl_pretrain}
\end{table}

\noindent\textbf{Analysis of sampling points.}
The number of sampling points is a key factor in obtaining dense observations.
Theoretically, as the sequence of input images increases, the richer the resulting map will be. In \circledbrown{4} and \circledbrown{5} of Table~\ref{tab:abl_pretrain}, increasing the number of sampling points from $4$ to $20$, Trans4Map with MiT-B2 benefits from the dense observations and obtains a gain of ${+}2.31\%$ in mIoU. Comparing \circledbrown{8} and \circledbrown{9}, the consistent improvement is achieved by Trans4Map with MiT-B4.

\noindent\textbf{Analysis of data modalities.}
To ablate the effect of data modalities, RGB and RGBD inputs are compared by using single-modal and dual-modal Trans4Map.
Compared between the RGBD baseline SMNet and \circledbrown{2} in Table~\ref{tab:abl_pretrain}, our dual-modal Trans4Map has a $3.5\%$ mIoU gain, while both are trained on SUN-RGBD.
When using $4$ sampling points, the dual-modal Trans4Map performs better than the single-modal one, as compared in \circledbrown{3} and \circledbrown{4}.
It indicates the effectiveness of our dual-modal Trans4Map in harvesting cross-stream complementary features for boosting performance.

\noindent\textbf{Analysis of model complexities.}
To inspect the trade-off between efficiency and performance, we analyze the model complexity by using three different backbones, \ie, MiT-B2, -B3 and -B4. They have respective $27.5$M, $47.4$M and $64.1$M parameters, yet all are more lightweight than the baseline SMNet with $83.9$M, as presented in Table.~\ref{tab:abl_backbone}. Comparing three backbones (\circledbrown{4}\circledbrown{6}\circledbrown{8}), larger models have better results, since more training data is available based on $4$ sampling points. In the case of $20$ sampling points (\circledbrown{5}\circledbrown{7}\circledbrown{9}), the three models achieve competitive results compared to the baseline model. This experiment demonstrates the effectiveness of our Trans4Map framework, achieving consistent improvements based on backbones of different sizes. 

\begin{figure*}[t]
    \centering
    \includegraphics[width=1.0\linewidth, keepaspectratio]{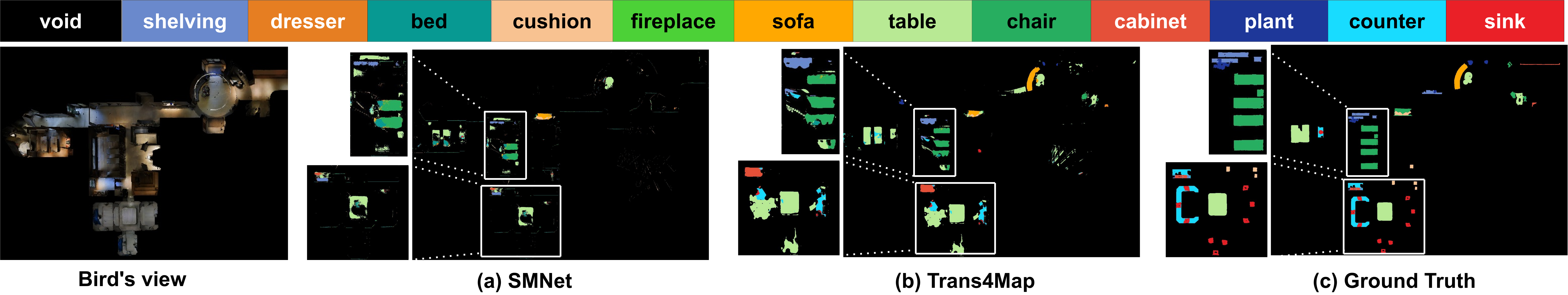}
    \vskip -2ex
    \caption{\textbf{Visualizations of the challenging case.} Zoom in for better view.} 
    \label{fig:vis_fail}
    \vskip -1ex
\end{figure*}

\subsection{Qualitative analysis}\label{sec4.5:vis}
\noindent\textbf{Semantic map visualizations.}
We visualize the semantic map results from the test set of the Matterport3D dataset, as shown in Fig.~\ref{fig:vis}. Thanks to the extracted non-local features and long-distance dependencies, Trans4Map has much better segmentation results. In the first scene in Fig.~\ref{fig:vis}, Trans4Map is better at segmenting the \emph{bed}. Further, Trans4Map is able to successfully classify the \emph{fireplace}, while the baseline model fails and predicts it as a \emph{cabinet}. In the second scene in Fig.~\ref{fig:vis}, Trans4Map delivers semantic mapping accurately, such as on \emph{cabinet} and \emph{chair} categories, while SMNet misclassifies them as \emph{tables}. Besides, SMNet yields incomplete \emph{chair} segmentation results.

\noindent\textbf{Challenging case analysis.}
The semantic mapping in the large-scale indoor scenes is still a challenging task. Fig.~\ref{fig:vis_fail} demonstrates that both models struggle with semantic mapping in such a large-scale scene. SMNet has difficulty segmenting the four \emph{chairs} completely as shown in Fig.~\ref{fig:vis_fail}~(a). Trans4Map performs slightly better, but there is still a large improvement space. 
All the six \emph{sinks} in the lower part are not predicted correctly. The reason is that the depth information of this scene is less reliable, while the whole scene covers about $2500$ square meters on the allocentric map. One potential solution is to obtain more observations to alleviate the error caused by the noisy depth measurements.

\section{Conclusion}
In this paper, we propose an end-to-end transformer-based framework, termed \emph{Trans4Map}, in order to revisit the egocentric-to-allocentric mapping from the front-view images to bird's-eye-view semantics. Based on the transformer-driven backbone and the \emph{Bidirectional Allocentric Memory (BAM)} updater, Tran4Map sets the new state of the art on both Matterport3D and Replica datasets, while using a more lightweight architecture and having fewer parameters as compared to the previous work. In the future, we will further explore domain adaptation methods to transfer the mapping models trained on the synthetic datasets to real-world scenes. Based on the constructed semantic map, the downstream tasks such as indoor navigation and path planning will also be interesting research directions.

\clearpage

\newpage
{\small
\bibliographystyle{ieee_fullname}
\bibliography{egbib}
}

\end{document}